\documentclass[conference]{IEEEtran}
\usepackage{cite}
\usepackage{graphicx}
\usepackage{multirow}
\usepackage{subcaption}
\usepackage{amsmath,amssymb,amsfonts}
\usepackage{graphicx}
\usepackage{textcomp}
\usepackage{algorithm}
\usepackage[noend]{algpseudocode}
\usepackage[normalem]{ulem}
\usepackage{fancyhdr}
\usepackage{dsfont}
\usepackage{comment}
\usepackage{enumitem}
\usepackage{url}
\usepackage[lmargin=0.7in, rmargin=0.7in, tmargin=0.7in, bmargin=1in]{geometry}
\usepackage{multirow}
\usepackage{rotating}
\usepackage[table,xcdraw]{xcolor}
\usepackage[normalem]{ulem}
\usepackage[framemethod=TikZ]{mdframed}
\usepackage[english]{babel}
\usepackage{amsthm}
\usepackage{hyperref}
\usepackage{booktabs}
\usepackage{xcolor}
\usepackage{colortbl}
\usepackage{array}
\usepackage{tikz}

\newmdtheoremenv[
leftmargin=3pt,
rightmargin=3pt,
backgroundcolor=gray!40,
roundcorner=4pt,
theoremtitlefont=\bfseries,
]{obsv}{Observation}

\renewcommand{\arraystretch}{1.2}

\def\BibTeX{{\rm B\kern-.05em{\sc i\kern-.025em b}\kern-.08em
    T\kern-.1667em\lower.7ex\hbox{E}\kern-.125emX}}

\title{Learning in Symbiosis: Heterogeneous Federated Learning with ANNs and SNNs}
 \author{\IEEEauthorblockN{Manh V. Nguyen\IEEEauthorrefmark{1},
   Liang Zhao\IEEEauthorrefmark{1},
   Bobin Deng\IEEEauthorrefmark{1},
   Xinyue Zhang\IEEEauthorrefmark{1},
   Shaoen Wu\IEEEauthorrefmark{1}}
   \IEEEauthorblockA{\IEEEauthorrefmark{1}College of Computing and Software Engineering,
   Kennesaw State University, Georgia, USA
   \\mnguy126@students.kennesaw.edu, \{lzhao10, bdeng2, xzhang48, swu10\}@kennesaw.edu}
 }

\begin{document}

\maketitle

\begin{abstract}
Spiking Neural Networks (SNNs), widely regarded as the third generation of neural networks, represent a promising neuromorphic computing paradigm with the potential to reduce the energy footprint of Artificial Intelligence (AI). 
Their energy efficiency makes SNNs particularly attractive for resource-constrained Edge AI devices, motivating growing research interest in Federated Learning (FL) with SNNs. 
Despite these advantages, integrating SNNs into existing FL systems remains challenging, as replacing all conventional Artificial Neural Network (ANN) devices with SNN-based alternatives is impractical. 
Instead, we envision a gradual transition toward neuromorphic Edge AI, where ANN and SNN devices coexist and collaboratively participate in FL.
To facilitate this transition, we propose Hybrid Neuromorphic Federated Learning (HNFL), a framework that enables joint training of heterogeneous ANN and SNN devices within an FL environment. 
Through experiments on CIFAR-10 using both VGG9 and ResNet architectures, we demonstrate the mutual benefits of collaborative ANN-SNN training. 
Specifically, HNFL achieves an average improvement of 16.85\% when incorporating SNN devices into ANN-based FL and 8.02\% when incorporating ANN devices into SNN-based FL. 
Furthermore, HNFL produces sparser spiking activity in the trained SNN models, with reductions of 12.4\% for SResNet and 22.4\% for SVGG9.
To the best of our knowledge, this is the first work to demonstrate that ANNs and SNNs can coexist and mutually benefit from collaborative training in FL, providing a practical path toward the gradual integration of neuromorphic computing into existing Edge AI systems.

\end{abstract}

\section{Introduction}
\label{sec:intro}

As energy-efficiency in Edge AI become increasingly critical in resource-constraint devices under strict latency and power budgets~\cite{shi2016edge}, Spiking Neural Networks (SNNs) emerged as a potential solution thanks to their event-driven, neuromorphically designed nature~\cite{roy2019towards}. 
To address the need for security, Federated Learning (FL) has also garnered massive interest in Edge AI research, enabling collaborative model optimization without centralizing raw user data~\cite{mcmahan2017communication}.
Despite this convergence of interests in the field of FL-SNNs, studies of practical SNNs deployment in federated edge systems remains an under-explored research area. 

The arrival of neuromorphic hardware, such as Intel Loihi and IBM TrueNorth~\cite{merolla2014million, davies2018loihi}, has helped demonstrated the practicality and energy-efficiency potential of SNNs. 
Simultaneously, the maturing investigation into FL-SNNs informs us of a more robust and communication efficient edge learning system \cite{venkatesha2021federated, nguyen2024robustness}. 
Nonetheless, there are still hesitant in the adoption of SNNs into Edge AI industry-wise, this is partially due to the potentially significant commitment cost. 
Namely, the prospect of abandonning the working traditional edge systems in place of a massive hardware upgrades expense. 

Instead of a risky and disrupting approach to neuromorphic adoption, we speculate that a slow incorporation of neuromorphic devices into the existing traditional Edge AI systems to be a more effective one. 
To that end, an ecosystem where both ANNs and SNNs can cohabitate and collaborate with each others is needed if the integration is to be deemed feasible.
Unfortunately, limited attention has been paid to the study of FL training in a heterogeneous environemnt of traditional and neuromorphic models, much less, demonstrating the advantages of such systems (further survey in Section \ref{subsec:fedjoint}). 
To address research gap, we propose \textit{Hybrid Neuromorphic Federated Learning (HNFL)}, a federated optimization framework that enables ANN and SNN clients to train collaboratively by sharing backbone feature-extraction weights while maintaining partitioned normalization parameters. 

Our contributions in this paper are threefold:

\begin{itemize}
  \item We formally characterize the normalization incompatibility that arises during joint ANN–SNN training in Federated Learning (FL) and propose Hybrid Neuromorphic Federated Learning (HNFL), a framework that addresses this challenge through parameter partitioning.

  \item We conduct comprehensive experiments simulating FL environments comprising heterogeneous ANN and SNN devices. Our results demonstrate that collaborative training through HNFL improves the performance of both model types, achieving gains of up to 16.85\% compared to isolated training.

  \item We demonstrate that HNFL-trained SNN models exhibit sparser spiking activity than those trained in SNN-only FL environments, reducing the average firing rate by up to 22.4\% with the VGG9 architecture. This finding suggests additional potential for energy savings on neuromorphic hardware.

\end{itemize}

Our findings establish a foundation for collaborative ANN–SNN training in Edge AI and support a gradual integration strategy for neuromorphic intelligence, offering an alternative to the costly, large-scale replacement of existing traditional ANN hardware.

\section{Background \& Related Work}
\label{sec:related}

\subsection{Normalization in SNNs}\vspace{-2pt}

Wu et al. and Zheng et al. offers the earliest specific normalization techniques that speaks of the peculiarity of spiking neuronal activities that necessitates SNN-specific solution \cite{wu2019direct,zheng2021going}. 
The dynamics of the temporal-dimension continues to be the point of focus of later innovation in spiking normalization. 
In which, Kim et al. and Duan et al. \cite{kim2021revisiting,duan2022temporal} dedicates one set of normalization parameters per timestep, while Jiang et al. \cite{jiang2024tab} and Liu et al.\cite{liu2025enhancing} further analyzes the statistical behavior accross the timesteps.
The numerous investigation into spiking normalization has shown that, while ANN and SNN might intuitively share parameters that performs similar feature extraction, there is inherent mismatch in the normalization parameters that renders their joint training not-so-straightforward (further discussion in Section \ref{subsec:challenge}).

\vspace{-3pt}\subsection{Federated Learning with SNNs}\vspace{-2pt}
\label{subsec:fedjoint}
Federated learning (FL) is commonly built on FedAvg, which performs local SGD with periodic aggregation~\cite{mcmahan2017communication}, later work highlighted and tackled challenges from heterogeneity in model size and communication limits~\cite{diao2021heterofl,vepakomma2018split}, to heterogeneity (non-IID) of  data~\cite{li2020federated, li2020fedprox}.  
Due to its energy-efficiency of SNNs, investigation into its FL properties is a widely interested topic, Skatchkovsky et al.~\cite{skatchkovsky2020federated} showed that federated neuromorphic learning can approach centralized performance under communication constraints, and Venkatesha et al.~\cite{venkatesha2021federated} studied federated SNN training under non-IID settings. 
Our previous work \cite{nguyen2024robustness} demonstrate SNNs robustness under Byzantine attacks with low communication bandwidth consumption. 
However, these works have only explored FL training with SNN-only in juxtaposition with ANN-only environments.

\vspace{-3pt}\subsection{Joint ANN--SNN Training}\vspace{-2pt}
In terms of FL training both ANNs and SNNs simultaneously, Yu et al. ~\cite{yu2024heterogeneous} experimented with a fusion model, blind aggregation of all parameters across the CNN architecture. 
This approach completely neglect the nuance of data encoding, temporal dynamics, and normalization behavior of SNNs in comparison to ANNs, thus over idealized their similarity. 
Tao et al. \cite{tao5995698federf} understand this nuance and introduce a ranking system to determine the set of parameters that would be jointly aggregated between both ANN and SNN models. However, they only regard the first (input) layer as potentially valuable to share among both models, which reduce the amount of sharable parameters to miniscule, thus risk limiting the amount of knowledge that can both models can learn from one another.
To our best knowledge, our work is the first to highlight the misalignment problem in data encoding, temporal dynamics and normalization of joint ANN--SNN training, our proposed FL training scheme segregate the normalization and shared feature sets of parameters to address this problem.

\section{Hybrid Neuromorphic Federated Learning}
\label{sec:method}

In this section, we briefly outline the difficulty in joining the training of ANN and SNN devices together in Section \ref{subsec:challenge}. 
Then, in Section \ref{subsec:model_decomposition}, we describe our how proposed algorithm, dubbed Hybrid Neuromorphic Federated Learning (HNFL), and how it overcomes the outlined challenge. 
In Section \ref{subsec:federated_optimization}, we provide the pseudo-code and detailed explanation of the federated optimization process in HNFL.

\begin{figure}
    \centering
    \includegraphics[width=0.8\linewidth]{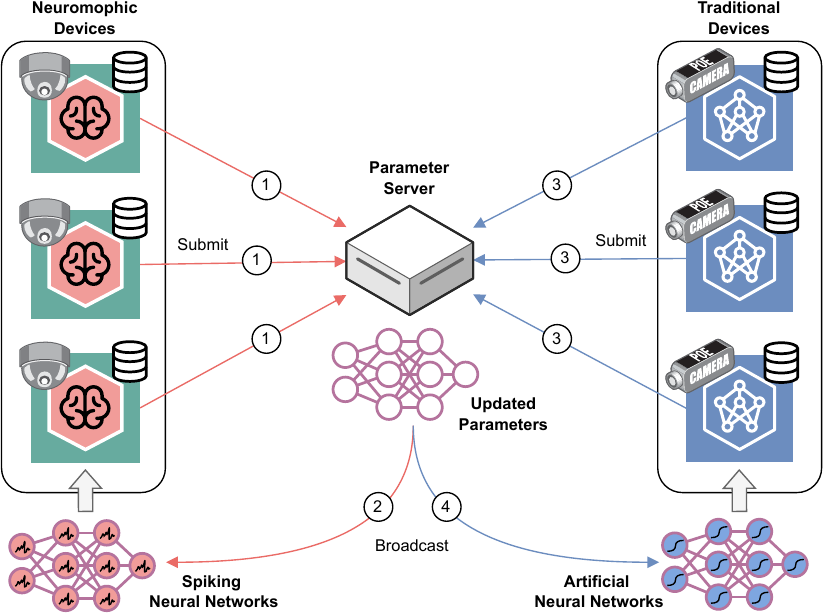}
    \caption{Hybrid Neuromorphic Federated Learning (HNFL) operates in a heterogeneous network comprising both conventional and neuromorphic devices. Each device collects its own modality-specific data and collaboratively contributes to training a unified ANN–SNN model.}
    \label{fig:concept}
\end{figure}

\subsection{The Challenge of Joint ANN--SNN Models Training}
\label{subsec:challenge}

Let us assume an FL ecosystem of mixed neuromorphic and traditional devices, and we are performing a joint training of both model types. For the ANN models, we assume the usage of original batch normalization (BN) \cite{ioffe2015batch}, in which the BN pass for the activation at layer $l$ is:
\begin{equation}
\mathrm{BN}^{(l)}(a^{(l)})
=
\gamma^{(l)}
\frac{a^{(l)} - \mu^{(l)}}{\sqrt{(\sigma^{(l)})^2 + \varepsilon}}
+
\beta^{(l)}.
\end{equation}
with $\gamma^{(l)}$ and $\beta^{(l)}$ as learnable affine parameters, $\mu^{(l)}$ and $\sigma^{(l)}$ as the mean and standard deviation; these are computed on each channel of the activation map $a^{(l)}$. 

For the SNN models, we implement batch norm through time (BNTT) \cite{kim2021revisiting},
in which, the forward pass on layer $l$ introduce the temporal dimension, i.e. normalizing the spiking output of each timestep requires a separated set of $\{\gamma_{(l)}, \beta_{(l)}, \mu_{(l)}, \sigma_{(l)}\}$: 
\begin{equation}
\text{BNTT}^{(l)}_t(s_t^{(l)})
=
\gamma_t^{(l)}
\frac{s_t^{(l)} - \mu_t^{(l)}}{\sqrt{(\sigma_t^{(l)})^2 + \varepsilon}}
+
\beta_t^{(l)}.
\end{equation}
This introduce an \textit{inherent mismatch of parameters between the ANN and SNN's normalization} as we naively perform joint ANN--SNN training by aligning each layer's parameters for aggregation. 
Say $\phi_{\mathrm{ann}}$ denotes ANN-specific, and $\phi_{\mathrm{snn}}$ denotes SNN-specific normalization parameters, we will have $\left|\phi_{\mathrm{snn}}\right| = T \times \left|\phi_{\mathrm{ann}}\right|$; with $T$ as the number of SNN timesteps. 

\subsection{The Hybrid Training Model}
\label{subsec:model_decomposition}

In this section, we provide a detailed description of our proposed HNFL algorithm, which overcome the Batchnorm-misalignment challenge (as outlined in the previous section) by maintaining separate sets of parameters each aggregated differently. 
Specifically, let us decompose the global model as follows: 
\[
\begin{aligned}
  &f(x; \Theta_\mathrm{global})
\\
  &\textrm{with: } \Theta_\mathrm{global} = \theta \cup \theta_{\mathrm{ann}} \cup \theta_{\mathrm{snn}}.
\end{aligned}
\]
where $\Theta_\mathrm{global}$ is the global parameter set, \(\theta\) denotes parameter set shared across both ANN and SNN models, these consist of convolutional layers, linear layers, and other feature extraction weights; $\theta_\mathrm{snn}$ is the parameter set exclusive to SNN, and $\theta_\mathrm{ann}$ is exclusive to ANN. 
To by pass the mismatch identified in Section \ref{subsec:challenge}, we separate the batch norm parameters of ANN and SNN, namely, $\theta_\mathrm{snn} = \phi_\mathrm{snn}$, and $\theta_\mathrm{ann} = \phi_\mathrm{ann}$.
The ANN and SNN models are then written as:
\[
\begin{aligned}
  &f_{\mathrm{ann}}(x) = h_{\mathrm{ann}}(x; \{\theta, \theta_{\mathrm{ann}}\})
\\
  &f_{\mathrm{snn}}(x_{[1,T]}) = h_{\mathrm{snn}}(x_{[1,T]}; \{\theta, \theta_{\mathrm{snn}}\})
\end{aligned}
\]

% This decomposition reflects the central idea of hybrid FL: the representation backbone is shared across all clients, while normalization remains architecture-specific. 
% In particular, \(\theta\) is updated by both ANN and SNN clients, \(\phi_{\mathrm{ann}}\) is updated only by ANN clients, and \(\phi_{\mathrm{snn}}\) is updated only by SNN clients.

Under this design, ANN and SNN clients share the same weights \(W^{(l)}, b^{(l)} \in \theta\) at layer \(l\), but different normalization parameters. 
Specifically, for an ANN client, the activation map at layer \(l\) is computed as:
\[
  a^{(l)} = \mathrm{ReLU}\!\left(
\mathrm{BN}_{\theta_{\mathrm{ann}}^{(l)}}\!\left(
W^{(l)} a^{(l-1)} + b^{(l)}
\right)\right).
\]
On the other hand, at timestep $t$, the SNN spiking activation at layer $l$ is computed according to the LIF model \cite{gerstner2014neuronal} as:
\[
s_{t}^{(l)}
=
\begin{cases}
1, & \tilde{U}_{t}^{(l)} \geq V_{\mathrm{th}}\\
0, & \tilde{U}_{t}^{(l)} < V_{\mathrm{th}}
\end{cases}
\]
with the pre-spike membrane potential computed as:
\[
\tilde{U}_{t}^{(l)}
=
U_{t-1}^{(l)}
+
\mathrm{BNTT}_{\theta_{\mathrm{snn}}^{(l),t}}
\left(
W^{(l)}s_{t}^{(l-1)} + b^{(l)}
\right)
\]
where $U_{t-1}^{(l)}$, whose formula is omitted for readability, is the post-spike membrane potential at the previous timestep, and $V_\mathrm{th}$ represents the spike threshold. 

\subsection{Federated Optimization}
\label{subsec:federated_optimization}

\begin{algorithm}[t]
\caption{Hybrid Neuromorphic Federated Learning} 
\label{alg:hnfl}
\textbf{Input:} $\mathcal{C}_{\mathrm{ann}}$, $\mathcal{C}_{\mathrm{snn}}$, $\{\mathcal{D}_i : i \in \mathcal{C}\}$, $R$, $\eta$, $\eta_{\mathrm{ann}}$, $\eta_{\mathrm{snn}}$ \\
\textbf{Output:} $\theta_R$, $\theta_{\mathrm{ann},R}$, $\theta_{\mathrm{snn},R}$
\begin{algorithmic}[1]
\State Initialize global parameters: $\theta_0$, $\theta_{\mathrm{ann},0}$, $\theta_{\mathrm{snn},0}$
\For{$r = 1$ to $R$}
    \State \textbf{Broadcast} $(\theta_r, \theta_{\mathrm{ann},r})$ to all $i \in \mathcal{C}_{\mathrm{ann}}$
    \State \textbf{Broadcast} $(\theta_r, \theta_{\mathrm{snn},r})$ to all $j \in \mathcal{C}_{\mathrm{snn}}$
    \For{each client $i \in \mathcal{C}$ \textbf{in parallel}}
        \If{$i \in \mathcal{C}_{\mathrm{ann}}$}
            \State Forward pass $f_{\mathrm{ann}}(x; \theta_r, \theta_{\mathrm{ann},r})$ on $\mathcal{D}_i$
            \State Compute $g_i^\theta = \nabla_\theta \mathcal{L}_i$
            \State Compute $g_i^{\theta_{\mathrm{ann}}} = \nabla_{\theta_{\mathrm{ann}}} \mathcal{L}_i$
            \State Send $(g_i^\theta, g_i^{\theta_{\mathrm{ann}}})$ to server
        \ElsIf{$i \in \mathcal{C}_{\mathrm{snn}}$}
            \State Forward pass $f_{\mathrm{snn}}(x_{1:T}; \theta_r, \theta_{\mathrm{snn},r})$ on $\mathcal{D}_i$
            \State Compute $g_i^\theta = \nabla_\theta \mathcal{L}_i$
            \State Compute $g_i^{\theta_{\mathrm{snn}}} = \nabla_{\theta_{\mathrm{snn}}} \mathcal{L}_i$
            \State Send $(g_i^\theta, g_i^{\theta_{\mathrm{snn}}})$ to server
        \EndIf
    \EndFor
    \State $\theta_{r+1} \gets \theta_r - \eta \sum_{i \in \mathcal{C}} p_i g_i^\theta$
    \State $\theta_{\mathrm{ann},r+1} \gets \theta_{\mathrm{ann},r} - \eta_{\mathrm{ann}} \sum_{i \in \mathcal{C}_{\mathrm{ann}}} p_i g_i^{\theta_{\mathrm{ann}}}$
    \State $\theta_{\mathrm{snn},r+1} \gets \theta_{\mathrm{snn},r} - \eta_{\mathrm{snn}} \sum_{i \in \mathcal{C}_{\mathrm{snn}}} p_i g_i^{\theta_{\mathrm{snn}}}$
\EndFor
\State \textbf{return} $(\theta_R, \theta_{\mathrm{ann},R}, \theta_{\mathrm{snn},R})$
\end{algorithmic}
\end{algorithm}

Let the set of ANN and SNN clients satisfy, 
\[
\mathcal{C}_{\mathrm{ann}} \cup \mathcal{C}_{\mathrm{snn}} = \mathcal{C},
\]
where each client \(i \in \mathcal{C}\) holds private data sampled from a local distribution \(\mathcal{D}_i\).
The hybrid federated objective jointly optimizes the shared parameters \(\theta\) together with the architecture-specific normalization parameters \(\theta_{\mathrm{ann}}\) and \(\theta_{\mathrm{snn}}\):
\[
\begin{aligned}
\min_{\theta,\theta_{\mathrm{ann}},\theta_{\mathrm{snn}}} \Bigg(
& \sum_{i \in \mathcal{C}_{\mathrm{ann}}} p_i \, \mathbb{E}_{(x,y)\sim \mathcal{D}_i}
\big[\mathcal{L}(f_{\mathrm{ann}}(x), y)\big] \\
& + \sum_{j \in \mathcal{C}_{\mathrm{snn}}} p_j \, \mathbb{E}_{(x,y)\sim \mathcal{D}_j}
\big[\mathcal{L}(f_{\mathrm{snn}}(x_{1:T}), y)\big]
\Bigg),
\end{aligned}
\]
where \(p_i\) denotes the aggregation weight of client \(i\), and \(\mathcal{L}(\cdot,\cdot)\) denotes the task loss function. 

Algorithm~\ref{alg:hnfl} summarizes the resulting training procedure. 
At round \(r\), client \(i\) computes the gradients of its local loss \(\mathcal{L}_i\) with respect to the shared parameters.
At each communication round, the server broadcasts the shared backbone \(\theta\) together with the matching normalization block to each client type: ANN clients receive \((\theta,\phi_{\mathrm{ann}})\), whereas SNN clients receive \((\theta,\phi_{\mathrm{snn}})\). 
Each client then performs local training on its private dataset \(\mathcal{D}_i\) using its corresponding model, namely \(f_{\mathrm{ann}}\) for ANN clients and \(f_{\mathrm{snn}}\) for SNN clients. The resulting local gradients
(denoted as $g_i^{\theta}$, $g_i^{\theta_{\mathrm{ann}}}$, and $g_i^{\theta_{\mathrm{snn}}}$ respectively) of the shared feature, ANN-norm, and SNN-norm parameter sets are then transmitted to the FL server for aggregation.
The server-side FL aggregation is performed on each update partitioned using conventional FedAvg formula. In particular, gradients for \(\theta\) are combined over all participating clients, while gradients for \(\theta_{\mathrm{ann}}\) and \(\theta_{\mathrm{snn}}\) are aggregated only within \(\mathcal{C}_{\mathrm{ann}}\) and \(\mathcal{C}_{\mathrm{snn}}\), respectively. 

\section{Evaluation}
\label{sec:experiments}
In this section, we demonstrate the result of the proposed hybrid training models through various metrics: accuracy, memory, communication, and computation.

\subsection{Experiment Setup}
We evaluate federated training of artificial neural networks (ANNs) and spiking neural networks (SNNs) on CIFAR-10 using two model families: a VGG-style architecture (VGG9/SVGG9) and a CIFAR-ResNet variant (ResNetCifar/SResNetCifar). In all experiments, training is distributed across 20 clients, with the composition of ANN and SNN clients varied to study homogeneous and heterogeneous federated regimes, including fully ANN, fully SNN, and mixed-client settings. Each client performs a local update on a single mini-batch by computing the forward pass, cross-entropy loss, and backpropagated gradients on its local data. Client updates are aggregated at the server by averaging the shared feature gradients, after which the global model is updated.

For SNNs, we consider two input configurations: 1) standard input repeat over time steps, and 2) Poisson random spike generation \cite{eshraghian2023training}. Batch-normalization statistics rollback is enabled in all reported experiment configurations to stabilize federated updates across clients. Model training utilized cross entropy loss and AdamW optimizer with the learning rate of $0.001$ and weight decay $5e-4$. The best ANN and SNN checkpoints are retained based on test performance. For the ResNet-based experiments, we use an Spiking Element Wise Additive (SEW-ADD) residual block \cite{fang_deep_2021} and WRN architecture specialized for CIFAR10 with depth parameter \(n=4\) and width multiplier \(k=2\) \cite{zagoruyko_wide_2017}. All the experiments are executed in parallel across multiple GPUs using a quota-based scheduling framework.

\subsection{Benchmark Client Participation Settings}

\begin{table}[t]
\centering
\begin{subfigure}[t]{\columnwidth}
\centering
\renewcommand{\arraystretch}{1.15}
\setlength{\tabcolsep}{6pt}

\scalebox{0.76}{
\begin{tabular}{c|cccc|cccc}
\hline
\multirow{3}{*}{\shortstack{\textbf{ANN}\\\textbf{Clients}\\\textbf{Added}}}
& \multicolumn{4}{c|}{\textbf{SNN Accuracy (\%)}}
& \multicolumn{4}{c}{\textbf{ANN Accuracy (\%)}} \\
\cline{2-9}
& \multicolumn{2}{c|}{\textbf{SResNet}}
& \multicolumn{2}{c|}{\textbf{SVGG9}}
& \multicolumn{2}{c|}{\textbf{ResNet}}
& \multicolumn{2}{c}{\textbf{VGG9}} \\
\cline{2-9}
& Repeat & Poisson
& Repeat & Poisson
& Repeat & Poisson
& Repeat & Poisson \\
\hline
0   & 72.55 & 69.28 & 79.67 & 72.67 & --    & --    & --    & --    \\
2   & 72.88 & 64.98 & 80.87 & 76.55 & 67.41 & 56.60 & 79.88 & 77.18 \\
4   & 72.19 & 66.49 & 82.12 & 77.79 & 70.19 & 69.24 & 82.05 & 80.61 \\
6   & 74.95 & 69.73 & 82.22 & 77.74 & 74.14 & 71.26 & 82.27 & 80.73 \\
8   & 76.82 & 68.37 & 82.87 & 77.78 & 77.35 & 70.69 & 83.74 & 81.33 \\
10  & 77.57 & 70.16 & 83.44 & 78.46 & 79.01 & 73.45 & 83.79 & 83.17 \\
\hline
\end{tabular}
}

\caption{Gradually increasing the number of incorporating ANN clients in an ecosystem full of SNN clients.}
\label{tab:ann_clients_increasing}
\end{subfigure}

\vspace{.2cm}

\begin{subfigure}[t]{\columnwidth}
\centering
\renewcommand{\arraystretch}{1.15}
\setlength{\tabcolsep}{6pt}
\scalebox{0.76}{
\begin{tabular}{c|cccc|cccc}
\hline
\multirow{3}{*}{\shortstack{\textbf{SNN}\\\textbf{Clients}\\\textbf{Added}}}
& \multicolumn{4}{c|}{\textbf{SNN Accuracy (\%)}}
& \multicolumn{4}{c}{\textbf{ANN Accuracy (\%)}} \\
\cline{2-9}
& \multicolumn{2}{c|}{\textbf{SResNet}}
& \multicolumn{2}{c|}{\textbf{SVGG9}}
& \multicolumn{2}{c|}{\textbf{ResNet}}
& \multicolumn{2}{c}{\textbf{VGG9}} \\
\cline{2-9}
& Repeat & Poisson
& Repeat & Poisson
& Repeat & Poisson
& Repeat & Poisson \\
\hline
0   & --    & --    & --    & --    & 74.12 & 73.17 & 80.49 & 80.82 \\
2   & 69.97 & 62.51 & 79.96 & 70.44 & 76.02 & 70.97 & 83.26 & 80.43 \\
4   & 69.04 & 62.72 & 81.20 & 73.81 & 73.06 & 67.25 & 83.28 & 82.08 \\
6   & 74.29 & 67.63 & 81.56 & 76.10 & 75.94 & 71.57 & 83.34 & 83.07 \\
8   & 74.44 & 65.71 & 83.03 & 78.84 & 76.46 & 69.48 & 83.73 & 83.66 \\
10  & 77.57 & 70.16 & 83.44 & 78.46 & 79.01 & 73.45 & 83.79 & 83.17 \\
\hline
\end{tabular}
}

\caption{Gradually increasing the number of incorporating SNN clients in an ecosystem full of ANN clients.}
\label{tab:snn_clients_increasing}
\end{subfigure}

\caption{Evaluating the gradual incorporation strategy with HNFL in different settings of neuromorphic and traditional FL devices participations.}
\label{tab:client_mixture_results}

\end{table}

To evaluate the practicality of HNFL under rolling device upgrades, we conduct a client-mixing study on CIFAR-10 with $|\mathcal{C}|=20$ clients, split into $|\mathcal{C}_{\mathrm{ann}}|=10$ ANN clients and $|\mathcal{C}_{\mathrm{snn}}|=10$ SNN clients. Across all settings, this partition is fixed while the number of active participants, $N_{\mathrm{ann}}$ and $N_{\mathrm{snn}}$, is varied. We report results for ResNet and VGG9, with SNNs using either repeated input at $T=4$ or Poisson spike encoding at $T=10$. Table~\ref{tab:client_mixture_results} contains two complementary scenarios: in Table~\ref{tab:ann_clients_increasing}, we fix $N_{\mathrm{snn}}=10$ and increase $N_{\mathrm{ann}}$ from $0$ to $10$, simulating the gradual inclusion of ANN devices into an SNN-based ecosystem; in Table~\ref{tab:snn_clients_increasing}, we fix $N_{\mathrm{ann}}=10$ and increase $N_{\mathrm{snn}}$ from $0$ to $10$, simulating the progressive deployment of SNN devices into an ANN-dominated federation. In both cases, we report the final ANN and SNN accuracies to assess how hybrid participation affects model performance.

The results show a consistent benefit from increasing participation of the complementary client type. In Table~\ref{tab:ann_clients_increasing}, progressively adding ANN clients to a federation with 10 SNN clients improves the SNN model across both architectures and input styles. For ResNet, SNN accuracy rises from $72.55\%$ to $77.57\%$ under repeated input (+$5.02$ points) and from $69.28\%$ to $70.16\%$ under Poisson input (+$0.88$). For VGG9, it increases from $79.67\%$ to $83.44\%$ under repeated input (+$3.77$) and from $72.67\%$ to $78.46\%$ under Poisson input (+$5.79$). The ANN model in the same table also improves as ANN participation grows: for ResNet, accuracy increases from $67.41\%$ to $79.01\%$ under repeated input (+$11.60$) and from $56.60\%$ to $73.45\%$ under Poisson input (+$16.85$); for VGG9, it rises from $79.88\%$ to $83.79\%$ under repeated input (+$3.91$) and from $77.18\%$ to $83.17\%$ under Poisson input (+$5.99$). Since no ANN clients are present at $N_{\mathrm{ann}}=0$, ANN accuracy is not applicable there, but once ANN clients are introduced, performance generally increases steadily. 

Table~\ref{tab:snn_clients_increasing} shows the reverse setting, starting from full ANN participation and gradually adding SNN clients, and again both model families benefit. For the SNN model, the gains are substantial: in ResNet, accuracy improves from $69.97\%$ to $77.57\%$ under repeated input (+$7.60$) and from $62.51\%$ to $70.16\%$ under Poisson input (+$7.65$); in VGG9, it rises from $79.96\%$ to $83.44\%$ under repeated input (+$3.48$) and from $70.44\%$ to $78.46\%$ under Poisson input (+$8.02$). The ANN model also benefits, though more moderately in some cases because it already starts from a strong baseline with 10 ANN clients: for ResNet, accuracy increases from $74.12\%$ to $79.01\%$ under repeated input (+$4.89$) and from $73.17\%$ to $73.45\%$ under Poisson input (+$0.28$); for VGG9, it rises from $80.49\%$ to $83.79\%$ under repeated input (+$3.30$) and from $80.82\%$ to $83.17\%$ under Poisson input (+$2.35$), despite some fluctuation at intermediate client counts. 

Overall, these results highlight the practical feasibility of HNFL for mixed-device federated ecosystems, showing that greater participation from either client type consistently benefits the hybrid training process. 
% Even when a federation is already fully populated by one modality, introducing more clients of the complementary type yields clear gains. For example, adding ANN clients to an SNN-only federation improves SNN accuracy by as much as $5.79$ percentage points in VGG9 with Poisson input and raises ANN accuracy by up to $16.85$ points in ResNet with Poisson input. 
% Conversely, gradually incorporating SNN clients into an ANN federation improves SNN accuracy by up to $8.02$ points in VGG9 with Poisson input, while still increasing ANN accuracy by as much as $4.89$ points in ResNet with repeated input. These substantial improvements show that neuromorphic hardware can be integrated incrementally without disrupting training, while also delivering measurable performance benefits to both model families.

\subsection{Benchmark Communication Efficiency}
\begin{figure}
    \centering
    \includegraphics[scale=0.4]{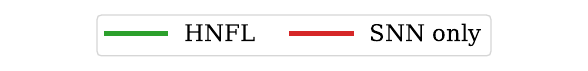}
    \\
    \begin{subfigure}{0.49\linewidth}
        \centering
        \includegraphics[width=\linewidth]{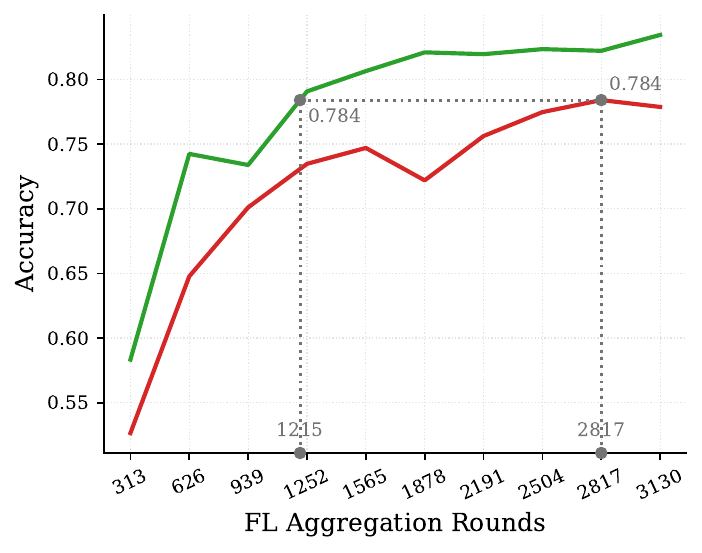}
        \caption{SVGG9 - Input: Repeat}
        \label{fig:total_agg_rounds_vgg_repeat}
    \end{subfigure}
    \begin{subfigure}{0.49\linewidth}
        \centering
        \includegraphics[width=\linewidth]{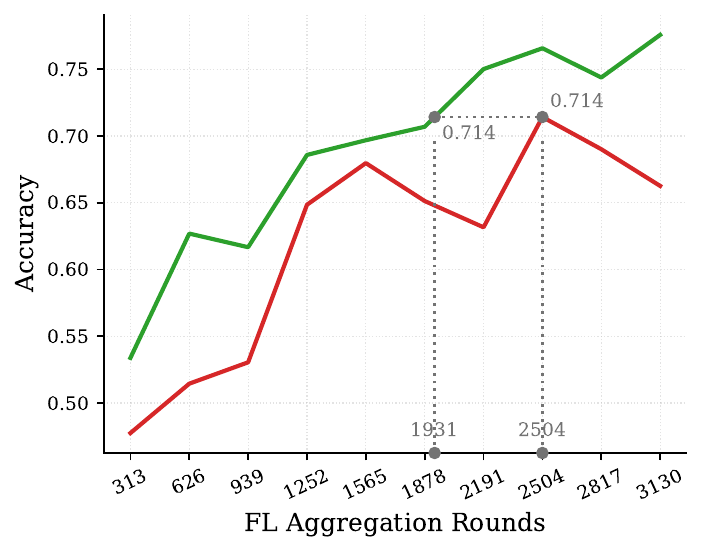}
        \caption{SResNet - Input: Repeat}
        \label{fig:total_agg_rounds_resnet_repeat}
    \end{subfigure}
    \begin{subfigure}{0.49\linewidth}
        \centering
        \includegraphics[width=\linewidth]{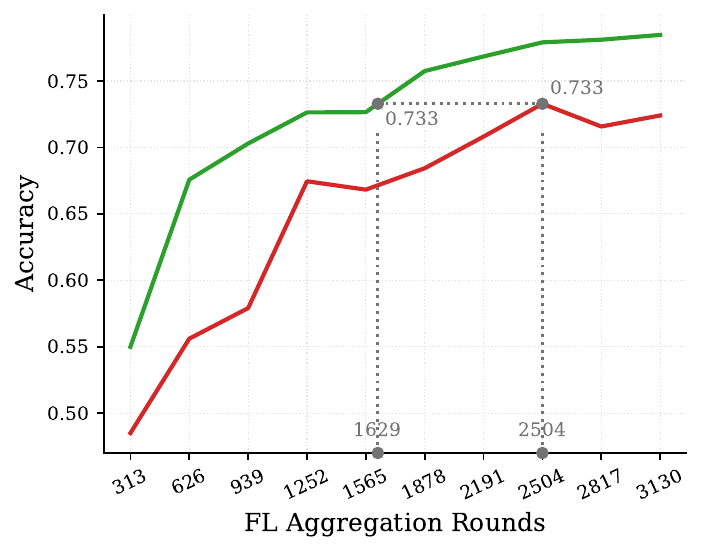}
        \caption{SVGG9 - Input: Poisson}
        \label{fig:total_agg_rounds_vgg_poisson}
    \end{subfigure}
    \begin{subfigure}{0.49\linewidth}
        \centering
        \includegraphics[width=\linewidth]{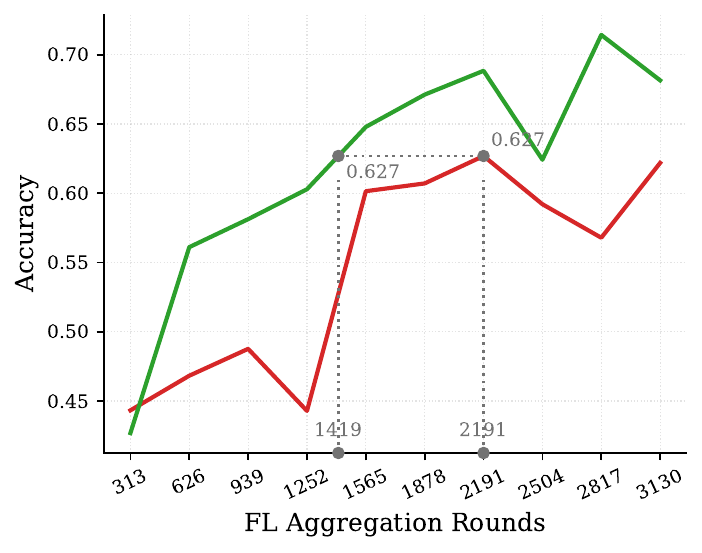}
        \caption{SResNet - Input: Poisson}
        \label{fig:total_agg_rounds_resnet_poisson}
    \end{subfigure}
    \caption{Communication efficiency of our propsed HNFL compared to pure FL-SNNs with all neuromorphic devices.}
    \label{fig:total_agg_rounds}
    \vspace{-1em}
\end{figure}

To further assess the practicality of HNFL, we evaluate its communication efficiency under the same client composition as above, with $|\mathcal{C}_{\mathrm{ann}}|=10$ and $|\mathcal{C}_{\mathrm{snn}}|=10$, by comparing: i) HNFL with full participation from both groups, and ii) an SNN-only setting with only the $10$ SNN clients active. Figure~\ref{fig:total_agg_rounds} plots SNN accuracy versus global aggregation rounds for VGG9 and ResNet under repeated and Poisson input encoding. For each setting, let $a_{\mathrm{SNNOnly}}^{\max}$ denote the maximum SNN accuracy achieved by the SNN-only baseline, and let $r_{\mathrm{SNNOnly}}$ and $r_{\mathrm{HNFL}}$ be the corresponding aggregation rounds at which the SNN-only and HNFL curves attain $a_{\mathrm{SNNOnly}}^{\max}$, respectively, with $r_{\mathrm{HNFL}}$ obtained by horizontal projection onto the HNFL curve. This allows us to compare the communication cost needed to reach the same target accuracy. In all four cases, we observe $r_{\mathrm{HNFL}} < r_{\mathrm{SNNOnly}}$, showing that HNFL achieves the SNN-only target accuracy with fewer global aggregation rounds.

For repeated input encoding, HNFL consistently reaches the best SNN-only accuracy in fewer aggregation rounds. 
In Figure~\ref{fig:total_agg_rounds_vgg_repeat}, the SNN-only VGG9 baseline attains $a_{\mathrm{SNNOnly}}^{\max}=0.784$ at $r_{\mathrm{SNNOnly}}=2817$, whereas HNFL reaches the same level at $r_{\mathrm{HNFL}}\approx1215$, reducing communication by $1602$ rounds ($56.9\%$), or equivalently requiring only $1/2.32$ of the rounds. 
The gain is already visible early in training, e.g., at round $626$, HNFL achieves $0.7424$ versus $0.6477$ for SNN-only, and after round $1252$ the HNFL curve stays near or above the final SNN-only peak. 
A similar trend appears for ResNet in Figure~\ref{fig:total_agg_rounds_resnet_repeat}: the SNN-only baseline reaches $a_{\mathrm{SNNOnly}}^{\max}=0.7142$ at $r_{\mathrm{SNNOnly}}=2504$, while HNFL reaches the same accuracy at $r_{\mathrm{HNFL}}\approx1931$, saving $573$ rounds ($22.9\%$), or about $1.30\times$ fewer rounds. 
Although the margin is smaller than for VGG9, HNFL remains ahead for most of training.

The same communication advantage is observed under Poisson encoding. 
In Figure~\ref{fig:total_agg_rounds_vgg_poisson}, the SNN-only VGG9 model reaches $a_{\mathrm{SNNOnly}}^{\max}=0.7328$ at $r_{\mathrm{SNNOnly}}=2504$, while HNFL reaches this target at $r_{\mathrm{HNFL}}\approx1629$, saving $875$ rounds ($34.9\%$), or about $1.54\times$ fewer rounds. 
Likewise, in Figure~\ref{fig:total_agg_rounds_resnet_poisson}, the SNN-only ResNet baseline achieves $a_{\mathrm{SNNOnly}}^{\max}=0.6269$ at $r_{\mathrm{SNNOnly}}=2191$, whereas HNFL reaches the same accuracy at $r_{\mathrm{HNFL}}\approx1419$, reducing communication by $772$ rounds ($35.2\%$), again about $1.54\times$ fewer rounds. 
In both Poisson settings, HNFL also exhibits a smoother and more stable improvement trajectory, while the SNN-only curves fluctuate more noticeably. 
Moreover, the final HNFL accuracies remain substantially higher than the best SNN-only results, reaching $0.7846$ for VGG9 and $0.7143$ for ResNet.

Overall, Fig. \ref{fig:total_agg_rounds} demonstrates a clear communication-efficiency advantage of HNFL, consistently reaching the best accuracy of the SNN-only baseline in fewer global aggregation rounds across all architectures and encoding schemes. 
Most notably, for repeated-encoding VGG9, HNFL reaches the SNN-only peak accuracy 2.32$\times$ faster, while the speedup remains substantial in the other settings at about 1.30$\times$ to 1.54$\times$ faster. 
These results strongly suggest that deploying SNN devices into a conventional ANN federated learning environment through HNFL is highly practical, as SNN clients can match or surpass SNN-only performance while converging significantly faster.

\subsection{Sampling Spike Activity}

\begin{figure}
    \centering
    \begin{subfigure}{\linewidth}
        \centering
        \includegraphics[width=0.9\linewidth]{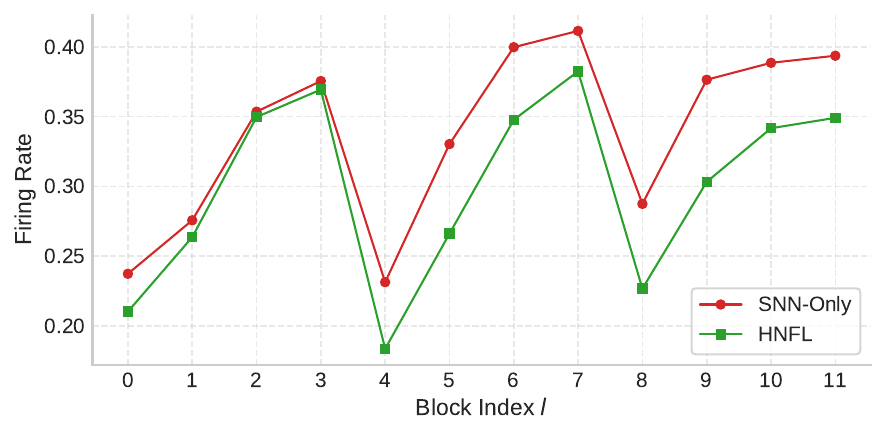}
        \caption{SResNet}
        \label{fig:firing_rate_resnet}
    \end{subfigure}
    \begin{subfigure}{\linewidth}
        \centering
        \includegraphics[width=0.9\linewidth]{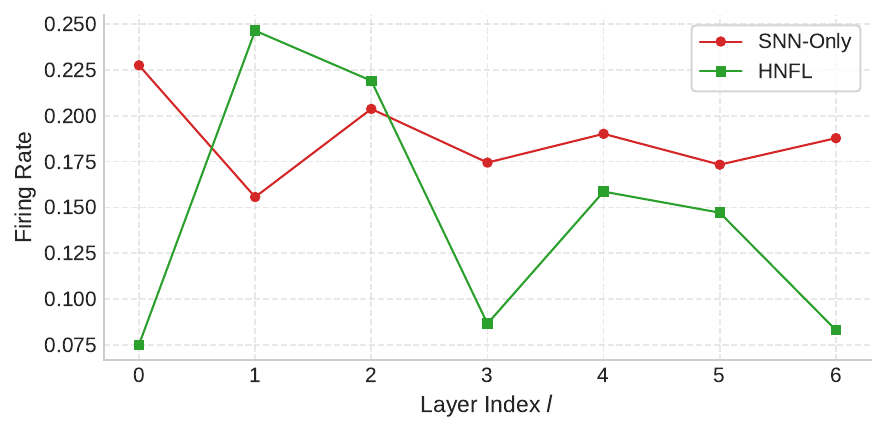}
        \caption{SVGG9}
        \label{fig:firing_rate_vgg9}
    \end{subfigure}
    \vspace{-1.2em}
    \caption{Spike rate by activation layers of SNN models trained by  SNN-Only and Hybrid methods.}
    \label{fig:firing_rate}
\end{figure}

Sparse activity in Spiking Neural Networks (SNNs) is a key property for neuromorphic efficiency, since fewer emitted spikes directly reduce the number of synaptic operations and thus the energy required for computation \cite{narduzzi2022optimizing}. In Fig.~\ref{fig:firing_rate}, we compare the layer-wise firing rates of SResNet and SVGG9 trained with SNN-Only and HNFL under repeating input with $T=4$ time steps. The training client composition is set to $|\mathcal{C}_{\mathord{ann}}| = |\mathcal{C}_{\mathord{snn}}| = 10$. For HNFL, we train full client participation, while for SNN-Only, the client configuration is $(N_{\mathord{snn}}, N_{\mathord{ann}}) = (10, 0)$.

For SResNet, HNFL yields a lower firing rate at every measured block, decreasing the average spike rate from $0.3418$ to $0.2995$, which corresponds to an absolute reduction of $0.0423$ and a relative reduction of approximately $12.4\%$. The largest reductions appear in deeper blocks, e.g., block~8 drops from $0.2875$ to $0.2267$ ($21.1\%$), block~9 from $0.3765$ to $0.3033$ ($19.4\%$), block~10 from $0.3886$ to $0.3417$ ($12.1\%$), and block~11 from $0.3937$ to $0.3492$ ($11.3\%$), while even earlier blocks show consistent improvements such as block~4, which decreases from $0.2313$ to $0.1836$ ($20.7\%$).
For SVGG9, the effect is more heterogeneous across layers, but HNFL still reduces the overall average firing rate from $0.1870$ to $0.1451$, corresponding to an absolute reduction of $0.0419$ or about $22.4\%$. In particular, substantial reductions are observed at layer~0, from $0.2275$ to $0.0750$ ($67.0\%$), layer~3, from $0.1744$ to $0.0865$ ($50.4\%$), and layer~6, from $0.1877$ to $0.0830$ ($55.8\%$), although HNFL slightly increases firing at layers~1 and~2. These results show that HNFL tends to produce SNN models with systematically sparser spiking activity than SNN-Only training, especially in later layers or blocks where event accumulation is often more costly, suggesting that the hybrid federated training scheme can improve not only model quality but also the energy efficiency of the resulting SNNs.

\section{Conclusions \& Future Work}
\label{sec:conclusion}
In this work, we investigate the question: \textit{``Can neuromorphic and traditional Edge AI devices mutually benefit from collaborative training in Federated Learning (FL)?''} Our findings provide a compelling affirmative answer. 
To enable such collaboration, we propose \textit{Hybrid Neuromorphic Federated Learning (HNFL)}, a framework designed to support joint ANN--SNN training in heterogeneous FL environments. 
We identify a key incompatibility between ANN and SNN normalization layers and address it through a parameter partitioning strategy that maintains three separate sets of parameters during training and aggregation.
More importantly, our experimental results demonstrate the mutual benefits of collaborative ANN--SNN training. 
Incorporating neuromorphic SNN devices into an ANN-only FL environment improves ANN classification accuracy by up to \(16.85\%\). 
Conversely, introducing conventional ANN devices into SNN-only FL improves SNN accuracy by \(8.02\%\). 
Furthermore, HNFL demonstrates advantages in communication efficiency, achieving comparable classification accuracy with, in some configurations, less than half the communication bandwidth. 
We also observe reduced spiking activity across the layers of HNFL-trained SNN models, suggesting further potential for reducing inference energy consumption on neuromorphic hardware.
Despite certain limitations, particularly the need to evaluate generalization under heterogeneous non-IID data distributions, our findings establish the feasibility and benefits of collaborative ANN--SNN training in FL. 
More broadly, this work advocates for the gradual integration of neuromorphic intelligence into existing Edge AI ecosystems, rather than the costly, large-scale replacement of conventional ANN devices.

\bibliographystyle{IEEEtran}
\bibliography{ref}

\end{document}